\documentclass[sigconf]{acmart}

\def\BibTeX{{\rm B\kern-.05em{\sc i\kern-.025em b}\kern-.08emT\kern-.1667em\lower.7ex\hbox{E}\kern-.125emX}}
    
\copyrightyear{2018}
\acmYear{2018}
\setcopyright{acmlicensed}
\acmConference[WSDM'21]{WSDM'21: ACM International Conference on Web Search and Data Mining}{March 08--12, 2021}{Jerusalem, Isreal}
\acmBooktitle{WSDM'21: ACM International Conference on Web Search and Data Mining, March 08--12, 20218, Jerusalem, Isreal}
\acmPrice{15.00}
\acmDOI{10.1145/1122445.1122456}
\acmISBN{978-1-4503-9999-9/18/06}

\usepackage{multirow}
\usepackage{caption}
\usepackage{subfig}

\newcommand{\nop}[1]{}

\begin{document}

%
% The "title" command has an optional parameter, allowing the author to define a "short title" to be used in page headers.
\title[Multi-Scale One-Class Recurrent Neural Networks]{Multi-Scale One-Class Recurrent Neural Networks \\for Discrete Event Sequence Anomaly Detection}

%
% The "author" command and its associated commands are used to define the authors and their affiliations.
% Of note is the shared affiliation of the first two authors, and the "authornote" and "authornotemark" commands
% used to denote shared contribution to the research.
\author{Zhiwei Wang}
\email{wangzh65@msu.edu}
\affiliation{%
  \institution{Michigan State University}
  \country{United States}
}

\author{Zhengzhang Chen}
\email{zchen@nec-labs.com}
\affiliation{%
  \institution{NEC LABORATORIES AMERICA, INC}
  \country{United States}
}

\author{Jingchao Ni}
\email{jni@nec-labs.com}
\affiliation{%
  \institution{NEC LABORATORIES AMERICA, INC}
  \country{United States}
}

\author{Hui Liu}
\email{liuhui7@msu.edu}
\affiliation{%
  \institution{Michigan State University}
  \country{United States}
}

\author{Haifeng Chen}
\email{huayi1983@gmail.com}
\affiliation{%
  \institution{NEC LABORATORIES AMERICA, INC}
  \country{United States}
}

\author{Jiliang Tang}
\email{tangjili@msu.edu}
\affiliation{%
  \institution{Michigan State University}
  \country{United States}
}

%
% By default, the full list of authors will be used in the page headers. Often, this list is too long, and will overlap
% other information printed in the page headers. This command allows the author to define a more concise list
% of authors' names for this purpose.

%
% The abstract is a short summary of the work to be presented in the article.
\begin{abstract}
Discrete event sequences are ubiquitous, such as an ordered event series of process interactions in Information and Communication Technology systems. Recent years have witnessed increasing efforts in detecting anomalies with discrete event sequences. However, it still remains an extremely difficult task due to several intrinsic challenges including data imbalance issue, discrete property of the events, and sequential nature of the data. To address these challenges, in this paper, we propose \textbf{OC4Seq}, a multi-scale one-class recurrent neural network for detecting anomalies in discrete event sequences. Specifically, \textbf{OC4Seq} integrates the anomaly detection objective with recurrent neural networks (RNNs) to embed the discrete event sequences into latent spaces, where anomalies can be easily detected. In addition, given that an anomalous sequence could be caused by either individual events, subsequences of events, or the whole sequence, we design a multi-scale RNN framework to capture different levels of sequential patterns simultaneously. Experimental results on three benchmark datasets show that \textbf{OC4Seq} consistently outperforms various representative baselines by a large margin.\nop{Extensive experiments on one synthetic and two benchmark datasets demonstrate the effectiveness of \textbf{OC4Seq}. Experimental results show that \textbf{OC4Seq} consistently outperforms various representative baselines by a large margin.} Moreover, through both quantitative and qualitative analysis, the importance of capturing multi-scale sequential patterns for event anomaly detection is verified. 
\end{abstract}



%
% Keywords. The author(s) should pick words that accurately describe the work being
% presented. Separate the keywords with commas.

%
% A "teaser" image appears between the author and affiliation information and the body 
% of the document, and typically spans the page. 

%
% This command processes the author and affiliation and title information and builds
% the first part of the formatted document.
\maketitle

\section{Introduction}

Nowadays, Information and Communication Technology (ICT) has permeated every aspect of our daily life and played a crucial role in society than ever. While ICT systems have brought unprecedented convenience, when in abnormal states caused by malicious attackers, they could also lead to ramifications including severe loss of economy and social wellbeing~\cite{gandhi2011dimensions,watkins2014impact,kopp2017cyber,byres2004myths,lagazio2014multi}. Therefore, it is vital to timely and accurately detect abnormal states of ICT systems such that the loss can be mitigated. Fortunately, with the ubiquitous sensors and networks, ICT systems have generated a large amount of monitoring data~\cite{malhotra2016lstm,ahmad2017unsupervised,zhou2017anomaly,du2017deeplog}. Such data contains rich information and provides us with unprecedented opportunities to understand complex states of ICT systems.

\begin{figure}
  \centering
  \includegraphics[scale=0.65]{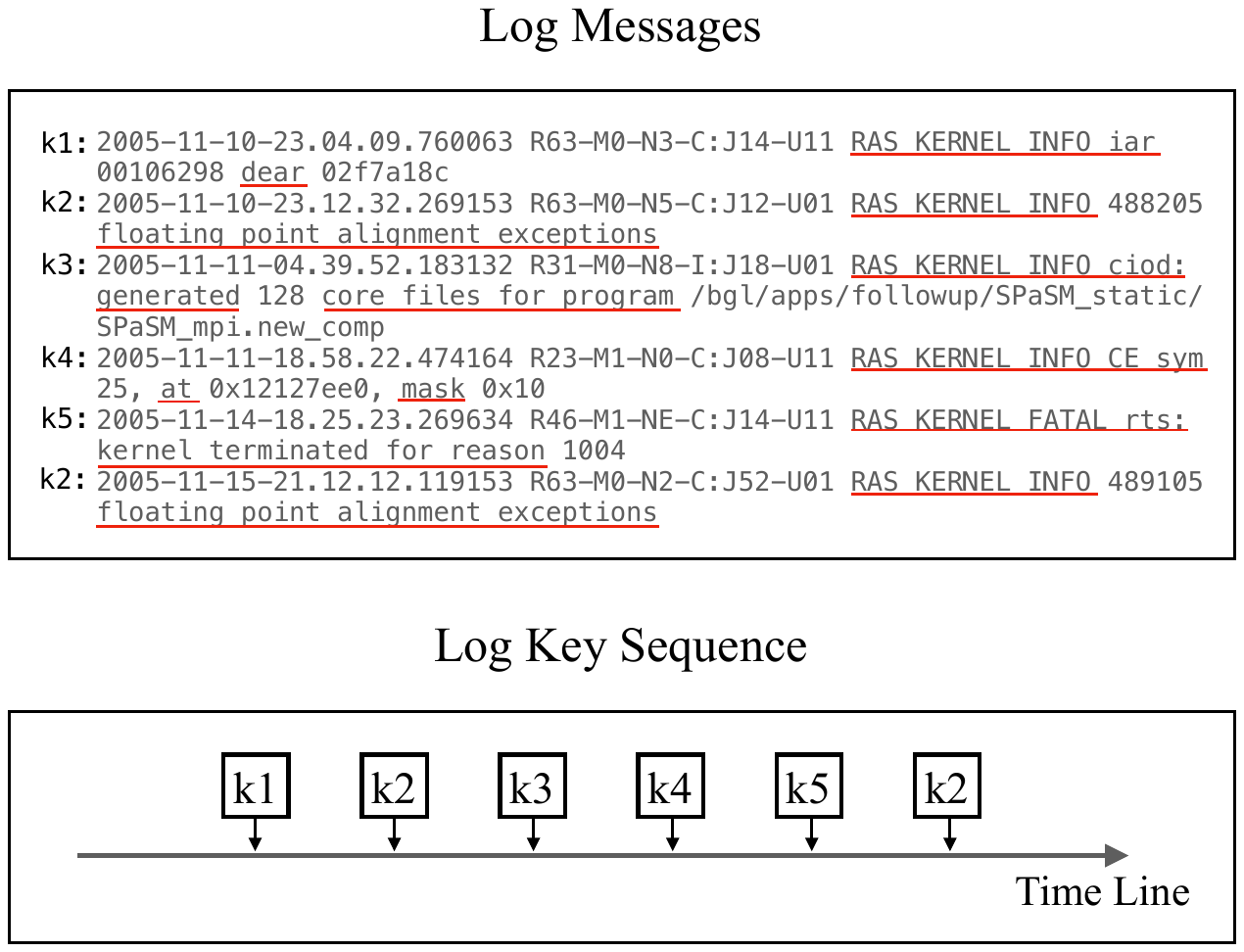}
  \caption{An illustrative example of BGL log messages and the corresponding log key sequence. Each log message contains a predefined log key that is underscored by red lines.}
  \label{fig:intro}
\end{figure}

One type of the most important monitoring data is discrete event sequence. A discrete event sequence is defined as an ordered series of events, where each event is a discrete symbol belonging to a finite alphabet~\cite{chandola2010anomaly}. The discrete event sequences can be seen everywhere such as control commands of machine systems, logs of a computer program, transactions of customer purchases, and DNA sequences in genetics. Due to the rich information they provide, they have been a valuable source for anomaly detection adopted by both academic research and industry practice~\cite{budalakoti2008anomaly, budalakoti2006anomaly, chandola2010anomaly,florez2005efficient,du2017deeplog}. For example, system logs that record the detailed messages of run time information by the modern computer systems are extensively used for anomaly detection~\cite{lou2010mining,du2017deeplog,xu2009detecting,oliner2007supercomputers}. Each log message can be roughly considered as consisting of a predefined constant print statement (also known as ``log key'' or ``message type'') and a specific parameter (also known as ``variable''). When the log keys are arranged chronologically according to the recording time, they form a discrete event sequence that can reflect the underlying system state. Figure~\ref{fig:intro} shows an illustrative example of logs from a BlueGene/L supercomputer system (BGL). In this example,\nop{there are} six log messages\nop{that} are generated by five corresponding predefined statements (log keys). These log keys form a discrete event sequence. When the system in an abnormal state, the resulted discrete event sequences will deviate from the normal patterns. For instance, if ``k2'' always comes after ``k1'' in a normal state, then the log key sequence (shown in Figure~\ref{fig:intro}) may indicate the abnormal state of the system because it shows that ``k2'' comes directly after ``k5'', which is unlikely to happen in the normal state.  In this work,  we refer to discrete event sequences generated by normal and abnormal system states as normal and abnormal sequences, respectively.

Despite that detecting anomaly for discrete event sequences has attracted much attention~\cite{marchetti2017anomaly,mitchell2014survey,chandola2010anomaly,gupta2013outlier,oak2019malware,du2017deeplog}, it still remains an extremely difficult task due to several intrinsic challenges. The first challenge is from the data imbalance issue that is commonly seen in anomaly detection problems. As systems are in normal states in most of the time, abnormal sequences are very rare. This makes the data distributed very unequally in terms of normal and abnormal states. Thus, binary classification models that have achieved great success in the other problems become ineffective for anomaly detection. Moreover, in reality, we often do not have prior knowledge about the abnormal sequence that further exacerbates the difficulty. The second obstacle comes from the discrete property of events. Unlike continuous sequences where each event is real-valued and have physical meanings, the discrete event sequence consists of discrete symbols, making it hard to capture the relations of events over time. Finally, the sequential nature of the data makes the problem even more challenging. In order to determine whether a discrete event sequence is abnormal or not, it is essential to consider each individual event, subsequences of events, and the whole sequence simultaneously. This requires dedicated efforts to designing models that not only have strong capability to capture the sequential patterns but also are flexible to handle sequential patterns at different scales.

To address the aforementioned challenges, in this paper, we propose \textbf{OC4Seq}, a multi-scale one-class recurrent neural network framework for event sequence anomaly detection. It is proposed to directly integrate the anomaly detection objective with a specially designed deep sequence model that explicitly incorporates sequential patterns at different scales. The main contributions of this work are summarized as follows:
\begin{itemize}
    \item We identify the importance of multi-scale sequential patterns in anomaly detection for discrete event sequences empirically.
    \item We introduce a novel framework \textbf{OC4Seq} to explicitly capture multi-scale sequential patterns and directly optimize the deep sequence models with a one-class classification objective. \textbf{OC4Seq} can be trained in an end-to-end manner.
    \item We conduct extensive experiments on three datasets to consistently demonstrate that OC4Seq outperforms the state-of-the-art representative methods with a significant margin.
\end{itemize}

\section{Problem Statement}
\label{Sec:PS}

Before we formally define the problem of anomaly detection for discrete event sequences, we firstly introduce notations that will be used throughout the rest of the paper. Lower-case letters such as $i$ and $j$ are used to denote scalar variables and upper-case letters such as $N$ and $M$ represent scalar constants. Moreover, we use bold lower-case letters to denote vectors such as ${\bf v}$ and ${\bf x}$ and bold upper-case for matrices such as ${\bf W}$. In addition, the $i^{th}$ entry of a vector ${\bf v}$ is denoted as ${\bf v}(i)$. Similarly, ${\bf W}(i,j)$ indicates the entry at $i^{th}$ row and $j^{th}$ column of a matrix ${\bf W}$. In the rest of the paper, event and discrete event are used interchangeably. We use $(\cdots)$ to represent an event sequence and subscripts are used to index the events in the sequence such as $(x_1, x_2, x_3)$.  

Given an event set $\mathcal{E}$ that contains all possible discrete events, an event sequence $S^i$ is defined as $S^i=(e^i_1, e^i_2, \cdots,  e^i_{N^i})$, where $e^i_j \in \mathcal{E}$ and $N^i$ is the length of sequence $S^i$. Each event $e^i_j$ is represented by a categorical value, \textit{i.e.}, $e^i_j \in \mathcal{N}^{+}$. 

With the notations above, the anomaly detection for discrete event sequence problem under the {\it one-class setting} is formally defined as follows:

{\it Given a set of sequences $\mathcal{S} = \{S^1, S^2, \cdots, S^N\}$, where each sequence $S^i$ is normal, we aim to design a one-class classifier that is able to identify whether a new sequence $S$ is normal or not by capturing the underlying multi-scale sequential patterns in $\mathcal{S}$.}

\section{Preliminaries: One-Class Classifier}

In this section, we introduce preliminaries that lay a foundation for our proposed framework. A one-class classifier is a specially designed classifier that is trained with objects of a single class and can predict whether an object belongs to this class or not in the test stage.  One of the most widely used one-class classifiers is kernel-based such as One-Class Support Vector Machines
(OC-SVM)~\cite{scholkopf2001estimating} and Support Vector Data Description (SVDD)~\cite{tax2004support}. Both OC-SVM and SVDD are inspired by SVM that tries to maximize the margin between two classes. Next, we use SVDD as an example to illustrate traditional one-class classifiers. SVDD aims at finding a spherically shaped boundary around the given data in the kernel space. Concretely, let the center of the hypersphere be ${\bf c}$ and the radius be $R> 0$. The SVDD objective is defined as follows:
\begin{align}
\label{eq:svdd}
&\min_{R, {\bf c},{\bf \varepsilon}}  R^2 + C \sum_{i=1}^{n}\varepsilon_i \\ \nonumber
s.t.  &  \|\phi({\bf x}_i - {\bf c})\|^2 \leq R^2 + \varepsilon_i, \varepsilon_i \geq 0, \quad \forall i
\end{align}
\noindent where ${\bf x}_i$ is the feature vector of $i^{th}$ data and $\varepsilon_i \geq 0$ is a slack variable for ${\bf x}_i$ that is introduced to allow the possibility of outliers in the training set and hyperparameter $C$ controls the trade-off between errors $\varepsilon_i $ and the volume of the sphere. The objective defined in Equation~\ref{eq:svdd} is in primary form and similar to SVM, it is solved in the dual space by using Lagrange multipliers. For more details of the optimization, please refer to the original paper~~\cite{tax2004support}. Once the $R$ and ${\bf c}$ are determined, the points that are outside the sphere will be classified as other classes.

\begin{figure*}
  \centering
  \includegraphics[scale=0.65]{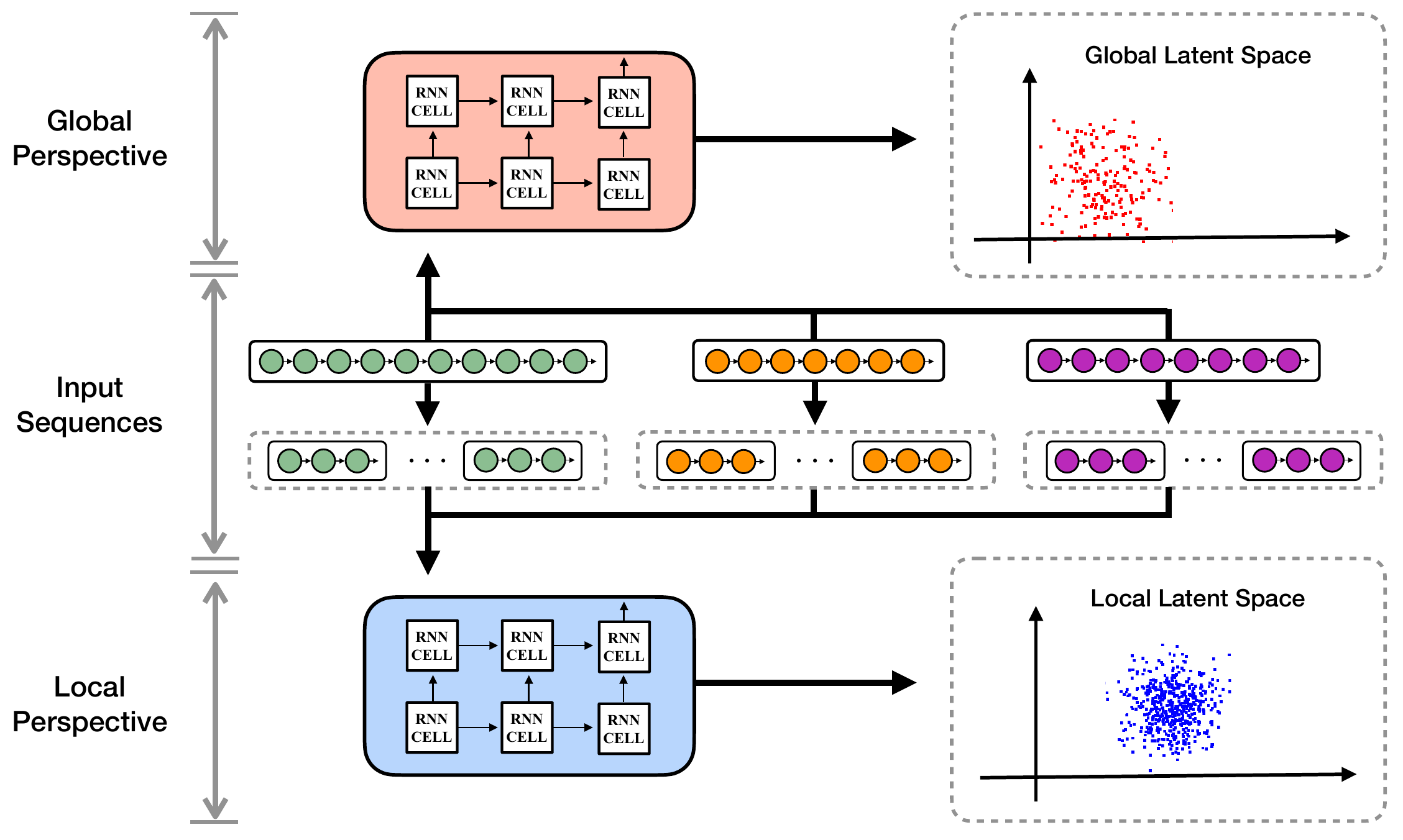}
  \caption{The overview of the proposed framework OC4Seq. It consists of two major components that learn the sequential information from global and local perspectives, respectively.}
  \label{fig:framework}
\end{figure*}

Recently, a deep neural network based one-class classifier called Deep SVDD was introduced in~\cite{ruff2018deep}. Inspired by SVDD, it tries to find a minimum hypersphere in the latent space. Unlike SVDD, which relies on kernel functions for feature transformation, Deep SVDD takes advantage of deep neural networks to learn data representations. It employs a quadratic loss for penalizing the distance of every data point representation to the center. The objective function has been proved with nice theoretical properties~\cite{ruff2018deep}. Once the neural networks are trained and center fixed, outliers will be detected similarly as SVDD.  \nop{Specifically, the simplified objective that employs a quadratic loss for penalizing the distance of every data point representation to the center is defined as:
\begin{align}
\min_{\mathcal{W}} \frac{1}{n} \sum_{i=1} \|\phi({\bf x}_i)-{\bf c}\|^2 + \frac{\lambda}{2}\|\mathcal{W}\|^2_F
\end{align}
\noindent where $\phi(\cdot)$ is the deep neural network whose parameter are $\mathcal{W}$ and $\|\|_F$ indicates the Frobenius norm. This objective function has been proved with nice theoretical properties~\cite{ruff2018deep}. Once the neural networks are trained and center fixed, outliers will be detected similarly as SVDD.  }

\section{The Proposed framework}

In this section, we introduce a multi-scale one-class recurrent neural network framework for event sequence anomaly detection. An overall of the proposed framework \textbf{OC4Seq} is shown in Figure~\ref{fig:framework}. It consists of two major components that focus on global and local information in the sequences, respectively. The details of each component are described in the next subsections.

\subsection{Learning Embeddings for Events}

The inputs of the framework are the sequences of events, where each event ${\bf e}_t$ is a one-hot vector and ${\bf e} (j)=1, {\bf e}(i)=0 \ \forall i \neq j$ if ${\bf e}_t$ is the $j^{th}$ type event of the set $\mathcal{E}$. In real-world scenarios, event space could be very large, \textit{i.e.}, there are tens of thousands of event types. This can lead ${\bf e}_t$ to be very high-dimensional and cause notorious learning issues such as sparsity and curse of dimension. In addition, one-hot vector representation makes an implicit assumption that events are independent with each other, which does not hold in most cases. Therefore, we design an embedding layer to embed events into a low-dimension space that can preserve relations between events. To do so, we introduce an embedding matrix ${\bf E} \in \mathbb{R}^{d^e \times |\mathcal{E}|}$, where $d^e$ is the dimension of the embedding space and $|\mathcal{E}|$ is the number of event types in $\mathcal{E}$. With this, the representation of ${\bf e}_t$ can be obtained as follows:
\begin{align}
{\bf x}_t = {\bf E}^T\cdot{\bf e}_t 
\end{align}
where ${\bf x}_t \in \mathbb{R}^{d^e} $ is the new low-dimensional dense representation vector for ${\bf e}_t$. After the embedding layer, the input sequences will be passed into the other components that will be introduced next.

\subsection{Anomaly Detection from Global Perspective}
To detect an anomalous sequence, it is important to learn an effective representation of the whole sequence in the latent space. To this end, we propose to integrate the widely-used Gated Recurrent Neural Networks (GRU)~\cite{cho2014learning} with one-class objective function. Specifically, given a normal sequence, \textit{i.e.}, $S^i = ({\bf x}^i_1, {\bf x}^i_2, \cdots, {\bf x}^i_{N^i})$, the GRU learns a representation of the sequence in a recursive manner. At the $t^{th}$ step, the GRU outputs a state vector ${\bf h}_t^i$, which is a linear interpolation between previous state ${\bf h}_{t-1}^i$ and a candidate state ${\bf \tilde{h}}_t^i$. Concretely:
\begin{align}
\label{eq:gru_1}
{\bf h}_t^i =  z_t^i \odot {\bf h}_{t-1}^i + (1-z_t^i) \odot {\bf \tilde{h}}_t^i
\end{align}
\noindent where $\odot$ is the element-wise multiplication. $z_t$ is the update gate which is introduced to control how much current state should be updated given the current information ${\bf x}_t^i$.  $z_t$ is calculated as:
\begin{align}
    z_t^i = \sigma({\bf W}_{z}{\bf x}_t^i + {\bf U}_{z}{\bf h}_{t-1}^i ) 
\end{align}
\noindent where ${\bf W}_{z}$ and ${\bf U}_{z}$ are the trainable parameters and $\sigma(\cdot)$ is a sigmoid function which is defined as follows:
\begin{align}
\sigma(x) = \frac{1}{1 + e^{-x}}
\end{align}
Moreover, the candidate state $ {\bf \tilde{h}_t}^i$ introduced in Equation~\ref{eq:gru_1} is computed as follows:
\begin{align}
    \label{eq:candidate_h}
    \tilde{h}_t^i=  g({\bf W}{\bf x}_t^i + {\bf U} (r_t^i \odot {\bf h}_{t-1}^i)) 
\end{align}
\noindent where $g(\cdot)$ is the tanh function that is defined:
\begin{align}
 g(x) = \frac{e^x - e^{-x}}{e^x + e^{-x}}
\end{align}
${\bf W}$ and ${\bf U}$ in Eq.~(\ref{eq:candidate_h}) are trainable parameters. $r_t$ is the reset gate. It is introduced to determine how much the candidate state should incorporate previous state. The reset gate is calculated as:
\begin{align}
    r_t^i =  \sigma({\bf W}_{r}{\bf x}_t^i + {\bf U}_{r}{\bf h}_{t-1}^i ) 
\end{align}

As the state vector ${\bf h}_{N^i}$ at the final step summarizes all the information in the previous steps, we regard it as the representation of the whole sequence. Please note that the GRU component can be replaced by any sequence learning models such as Long Short-Term Memory (LSTM)~\cite{hochreiter1997long}. In fact, we empirically found that the two have similar performance. Due to its structural simplicity, we choose GRU over LSTM. More details of the component analysis can be found in Section~\ref{Sec:experiment}.

Inspired by the intuition behind the Deep SVDD that all the normal data should be lie within a hypersphere of minimum volume in a latent space, we propose the following objective function to guide the training process:
\begin{align}
\mathcal{L}_{global} = \min_{{\bf \Theta}} \frac{1}{N} \sum_{i=1}^N \|{\bf h}_{N^i} - {\bf c}\|^2 + \lambda \|{\bf \Theta}\|_F^2
\end{align}

Here, ${\bf c}$ is a predefined center in the latent space and $n$ is the total number of sequences in the training set. The first term in the objective function employs a quadratic loss for penalizing the distance of every sequence representation to the center ${\bf c}$ and the second term is a regularizer controlled by the hyperparameter $\lambda$. Therefore, this objective will force the GRU model to map sequences to representation vectors that, on average, have the minimum distances to the center ${\bf c}$ in the latent space. 

Although the global GRU is effective to model the whole sequence, it might ignore vital information for event sequence anomaly detection because of the following reason: the abnormal property of a sequence can be caused by only a small abnormal subsequence or even a single abnormal event. However, when the sequence is long, the abnormal information could be overwhelmed by other normal subsequences during the representation learning procedure. This could lead to a very high false negative rate. 

\subsection{Anomaly Detection from Local Perspective}

In the previous subsection, we introduce to combine GRU and %Deep SVDD 
one-class classification objective to embed the normal sequences in a latent space where they lie within a small distance to a predefined center. However, local information that is vital for anomaly detection could be overwhelmed during this process. Thus, we design a subsequence learning component to detect the anomalies from the local perspective.

For a given event sequence, we construct subsequences of a fixed size $M$ with a sliding window. Therefore, each subsequence contains its unique local information, which plays an important role in determining whether the whole sequence is abnormal or not. To learn the representation of subsequence, we introduce the local GRU component that will model the sequential dependencies in every subsequence. To be concrete, given a subsequence ${\bf x}^i_{t-M+1}, {\bf x}^i_{t-M+2}, \cdots, {\bf x}^i_{t}$ of length $M$, the local GRU processes the events sequentially and outputs $M$ hidden states, the last of which is used as the representation of the local subsequence:
\begin{align}
{\bf h}^i_{t} = \text{GRU}({\bf x}^i_{t-M+1}, {\bf x}^i_{t-M+2}, \cdots, {\bf x}^i_{t})
\end{align}
Thus, for all subsequences in a sequence, the GRU will obtain a sequence of hidden representations that incorporate sequential dependencies in every local region as follows:
\begin{align}
{\bf h}^i_1, {\bf h}_2^i, \cdots, {\bf h}^i_{N^i-M} = LocalGRU({\bf x}^i_{1}, {\bf x}^i_{2}, \cdots, {\bf x}^i_{N^i})
\end{align}
\noindent where LocalGRU is the name for the second GRU model that processes each subsequence. For a normal event sequence, it is intuitive to assume that all of its subsequences are also normal. Thus, we further assume that all the local subsequences should be within a hypersphere in another latent space. To impose this assumption, we design the following objective function to guide the local sequence learning procedure:
\begin{align}
\mathcal{L}_{local} = \min_{{\bf \Theta}^L} \frac{1}{N} \sum_{i=1}^N \sum_{j=1}^{N^i-M} \|{\bf h}_{N^i_j} - {\bf c}^L\|^2 + \lambda \|{\bf \Theta}^L\|_F^2
\end{align}
Here, ${\bf c}^L$ is a predefined center of another hypersphere in the latent space and ${\bf \Theta}^L$ contains all the trainable parameters of LocalGRU. Similarly, the first term penalizes the average distance between all normal subsequences to the center ${\bf c}^L$ and the second term is a regularizer. 

\subsection{The Objective Function and Optimization Procedure}

In previous subsections, we have introduced components of OC4Seq to detect an abnormal event sequence from both global and local perspectives, respectively. In this subsection, we design an objective function to combine them together. Specifically, given the global and local loss function $\mathcal{L}_{global}$ and $\mathcal{L}_{local}$, the overall objective function of OC4Seq is defined as follows:
\begin{align}
\label{eq:obj}
\min_{{\bf \Theta}^L, {\bf \Theta}}\mathcal{L} = \mathcal{L}_{global} + \alpha \mathcal{L}_{local} 
\end{align}
where $\alpha$ is a hyper parameter that controls the contribution from local information in the sequence. This objective enables us to train the framework in an end-to-end manner. The specific optimization procedure is described next.

\noindent{\bf Optimization.} We use stochastic gradient descent (SGD) and its variants (\textit{e.g.}, Adam) to optimize the objective function defined in Eq.~(\ref{eq:obj}). Following previous work~\cite{ruff2018deep}, to accelerate the training process, the predefined centers ${\bf c}$ is computed as follows: given the untrained GRU, we firstly feed the sequences in the training set into it and obtain the sequence representation vectors. Then, we obtain an average vector by computing the mean value of all representation vectors and use it as ${\bf c}$. To obtain ${\bf c}^L$, a similar process is applied with untrained LocalGRU. Once ${\bf c}$ and ${\bf c}^L$ are obtained, they remain as constant vectors in the optimization process. The whole training process is done when the objective value converges.

\section{Experiment}
\label{Sec:experiment}
In this section, we evaluate the proposed framework OC4Seq on three datasets.
%In this section, we conduct extensive experiments to evaluate the proposed framework OC4Seq on one synthetic web logs and two real-world system log datasets. Next, we first describe the datasets and experimental settings. Then, we present the experimental results and observations. Finally, we conduct qualitative analysis to gain deep understandings on the proposed framework.

\subsection{Datasets}

\begin{table}
    \caption{The key statistics of \emph{RUBiS}, \emph{HDFS}, and \emph{BGL} datasets.}
        \label{tab:stats}
    \begin{center}  
    \begin{tabular}{lccc}
    \toprule
    Dataset & \# of normal  & \# of abnormal & \# of log keys\\
    \midrule                
    RUBiS & 11,677 & 1,000 & 24\\
    HDFS  & 558,221 & 16,838 & 28\\
    BGL & 9,543 & 985 & 1,540\\             
    \bottomrule
    \end{tabular}
    \end{center}
    %\vspace{-5mm}
\end{table}

\noindent{\bf RUBiS~\cite{amza2002specification}:} This dataset is a weblog dataset and was generated by an auction site prototype modeled after eBay.com~\cite{amza2002specification}. Specifically, each log message contains information related to a user web behavior including {\it user\_id}, {\it date}, {\it request\_info}, etc. Following previous work~\cite{yu2016cloudseer,zhao2016non}, we first parse each log message into a structured representation, which consists of a log key and a variable among others. Next, the log keys of the same user are collected following the time order that forms an event sequence. Thus, each log key sequence represents a user behavior session on a web server. 

Each abnormal sequence corresponds to an attack case. In total, there are $11,677$ normal sequences and $1,000$ abnormal sequences.

\noindent {\bf Hadoop Distributed File System (HDFS)~\cite{xu2009detecting}:} This dataset was generated by a Hadoop-based map-reduce cloud environment using benchmark workloads. It contains 11,175,629 log messages, of which 2.9\% are labeled as anomalies by Hadoop experts. Each log message involves a block ID, a timestamp, and state information. To make the comparison fair, we used the publicly available dataset\footnote{https://www.cs.utah.edu/~mind/papers/deeplog\_misc.html} processed by~\cite{du2017deeplog}. As described in~\cite{du2017deeplog}, the log messages are firstly parsed into structured text so that a log key is extracted from each log message. In addition, the log keys are sliced into sequences according to the associated block IDs. As a result, there are $558,221$ normal sequences and $16,838$ abnormal sequences.

\noindent {\bf BlueGene/L (BGL)~\cite{oliner2007supercomputers}:} This dataset contains $4,747,936$ log messages generated by a BlueGene/L supercomputer system with $131,072$ processors and $32,768$ GB memory at Lawrence Livermore National Labs. Each log message contains system information such as type, timestamp, nodes, content, etc. The log messages can be categorized into two types, \textit{i.e.}, non-alert and alert. The non-alert messages are labeled as normal and alert messages are labeled as abnormal. The log messages are firstly parsed by Drain~\cite{he2017drain} whose implementation\footnote{https://github.com/logpai/logparser} is open sourced by~\cite{zhu2019tools}. Following previous work~\cite{meng2019loganomaly}, the log keys are sliced using time sliding windows. A sequence is labeled as abnormal if it contains at least one abnormal message. After processing, there are $9,543$ normal sequences and $985$ abnormal sequences.

The statistics of the three datasets are summarized in Table~\ref{tab:stats}. As stated in Section~\ref{Sec:PS}, this work focuses on the one-class/unsupervised setting, where the training dataset does not contain any abnormal sequence. Therefore, each dataset is split into training, validation, and test sets by the following process. Firstly, we randomly sample the training data from the normal sequence set. Then, we separately split the remaining normal sequences and all the abnormal sequences into validation and test sets with the ratio 3/7 (validation/test). At last, we combine the two validation/test sets into one.

\begin{table*}
    \caption{The prediction performance comparison on \emph{RUBiS}, \emph{HDFS}, and \emph{BGL} dataset.}
        \label{tab:result}
    \begin{center}  
    \begin{tabular}{lccccccccc}
    \toprule
    \multirow{2}{*}{Method}  & \multicolumn{3}{c}{HDFS}  & \multicolumn{3}{c}{RUBiS} & \multicolumn{3}{c}{BGL}

    \\ \cmidrule(l){2-4}  \cmidrule(l){5-10}                
                            & F-1 score & Precision & Recall  & F-1 score & Precision & Recall & F-1 score & Precision & Recall  \\ 
    \midrule
    OC-SVM                  & 0.509 & 0.622 & 0.431 & 0.351 & 0.220 & 0.869 & 0.336 & 0.215  & 0.764\\ 
    PCA                     & 0.634 & 0.968 & 0.471 & 0.784 & 0.862 & 0.718 & 0.423 &0.269    &0.993\\ 
    Invariant Mining          & 0.943 & 0.893 & \textbf{1.000} & 0.912  & 0.841  & \textbf{0.996} & 0.428 &0.273  & \textbf{1.000}\\ 
    DeepLog                   & 0.941 & 0.952 & 0.930 & 0.935  & 0.885  & 0.992 & 0.326 &0.196 &0.980\\ 
    DeepSVDD                 & 0.872 & \textbf{0.964} & 0.796 & 0.970 & 0.974 & 0.966 & 0.410 &0.298 &0.659\\ 
    OC4Seq              & \textbf{0.976} & 0.955 & 0.998  & \textbf{0.985} & \textbf{0.987} & 0.983 &\textbf{0.747} &\textbf{0.704} &0.795\\ 
    \bottomrule
    \end{tabular}
    \end{center}
\end{table*}

\subsection{Baselines}
We compare OC4Seq with the following five anomaly detection baselines:

{\bf Principle Component Analysis (PCA)}~\cite{wold1987principal}. PCA is a classic unsupervised method that has been extensively used for a variety of problems. More recently, it becomes a popular method for anomaly detection~\cite{xu2009detecting}. Specifically, it firstly constructs a count matrix ${\bf M}$, where each row represents a sequence, each column denotes a log key, and each entry ${\bf M}(i,j)$ indicates the count of $j^{th}$ log key in the $i^{th}$ sequence. Next, PCA learns a transformed coordinate system, where the projection lengths of normal sequences are small while these of abnormal sequences are large. Although it has been shown that PCA can be effective in detecting anomalies especially in reducing false positives~\cite{du2017deeplog}, it ignores the sequential information, which could play an important role in event sequence anomaly detection. We use the open-sourced implementation\footnote{https://github.com/logpai/loglizer}~\cite{he2016experience}.

{\bf Invariant Mining (IM)}~\cite{lou2010mining}. IM is another popular unsupervised anomaly detection method. It is designed to automatically mine invariants in logs and assumes that the discovered invariants can capture the inherent linear characteristics of log flows. Similar to PCA, it firstly constructs a count matrix ${\bf M}$. Next, IM learns sparse, integer-valued invariants with physical meanings from ${\bf M}$. Finally, with the learned invariants, IM makes an invariant hypothesis. And the sequences that do not satisfy the hypothesis are detected as anomalies. As IM also relies on the $M$, it has similar drawbacks to PCA. The IM used in this work was implemented by~\cite{he2016experience}.

{\bf One-Class SVM (OC-SVM)}~\cite{scholkopf2001estimating}. OC-SVM is a very effective one-class classifier that has been extensively used for anomaly detection~\cite{li2003improving,wang2004anomaly,amer2013enhancing}. 

Specifically, it learns a kernel that maps the normal data into a latent space, where all the normal sequence clusters in a small region. Thus, a sequence that does not belong to the cluster is regarded as abnormal. To apply OC-SVM, we first need to extract features from each sequence. In this work, we tried two models to extract features: sequence auto-encoder~\cite{dai2015semi} and bag-of-words~\cite{zhang2010understanding}. As we empirically found the latter often has better performance, we choose bag-of-words as the feature extractor. The OC-SVM used in this work was implemented with the scikit-learn package\footnote{https://scikit-learn.org/}.

{\bf DeepLog}~\cite{du2017deeplog}. DeepLog is a state-of-the-art log anomaly detection method. This method is based on an LSTM model, which tries to capture the sequential dependencies in sequences. Specifically, by training with normal sequences, it learns to predict the next token given the previously seen tokens in a sequence. During the test stage, for each time step in a sequence, DeepLog will output a probability distribution over all the log keys. If any of the actual tokens are not in the top $k$ candidates, it will regard the sequence as abnormal. Compared to other baselines, this method can utilize sequential information and has demonstrated state-of-the-art performance in previous work. 

{\bf DeepSVDD}~\cite{ruff2018deep}. DeepSVDD is a general one-class classifier that builds on deep neural networks. It simultaneously learns a representation vector for each data and optimizes the anomaly detection objective directly. To make the comparison fair, we use the same RNN models in OC4Seq as its representation learning component. The implementation can be found at github\footnote{https://github.com/lukasruff/Deep-SVDD}.

\subsection{Experimental Settings}

{\bf Model Selection:} For all the methods with hyper-parameters, we use the validation set to select the best value and report the performance on the test set. For DeepLog, we follow the original paper's suggestion~\cite{du2017deeplog}. Specifically, both the $h$ and $g$ are selected from $\{8,9,10\}$, which denotes window size and candidate number, respectively. The number of layers is set to be $2$ and the number of LSTM hidden units is $64$. For OC4Seq, we use the same hyper-parameters as DeepLog and select $\alpha$ that controls the contribution of local subsequence from $\{0.01,0.1,1,10\}$. 

\noindent {\bf Implementation Details:} We implemented OC4Seq with Pytorch 1.5\footnote{https://pytorch.org/}. The model is trained using Adam~\cite{kingma2014adam} optimizer with the learning rate to be $0.01$. The mini-batch size is chosen to be $64$ and the model is trained for $100$ epochs on a single NVIDIA GEFORCE RTX 2080 card.

\noindent{\bf Evaluation Metrics:} To measure the model performance on anomaly detection, we choose the widely-used Precision, Recall, and F1 score as the evaluation metrics. They are defined as follows:
\begin{align}
 Precision &= \frac{TP}{TP + FP} \quad Recall  = \frac{TP}{TP + FN}\\ \nonumber
 %&  \\ \nonumber
 F1 &=\frac{2*Precision*Recall}{Precision + Recall} \nonumber
\end{align}
where TP (True Positives) denotes the number of true abnormal sequences that are detected by the model, FP (False Positives) measures the number of normal sequences that are regarded as anomalies, and FN (False Negatives) denotes the number of abnormal sequences the model fails to report. Thus, based on these definitions, there is a well-known trade-off between precision and recall. On the other hand, the F1 score considers the balance of the two and is often considered as a more comprehensive evaluation metric.

\begin{figure*}
\centering
\subfloat[Precision-Recall curves with different alpha.\label{fig:alpha}]{{\includegraphics[width=0.3\textwidth]{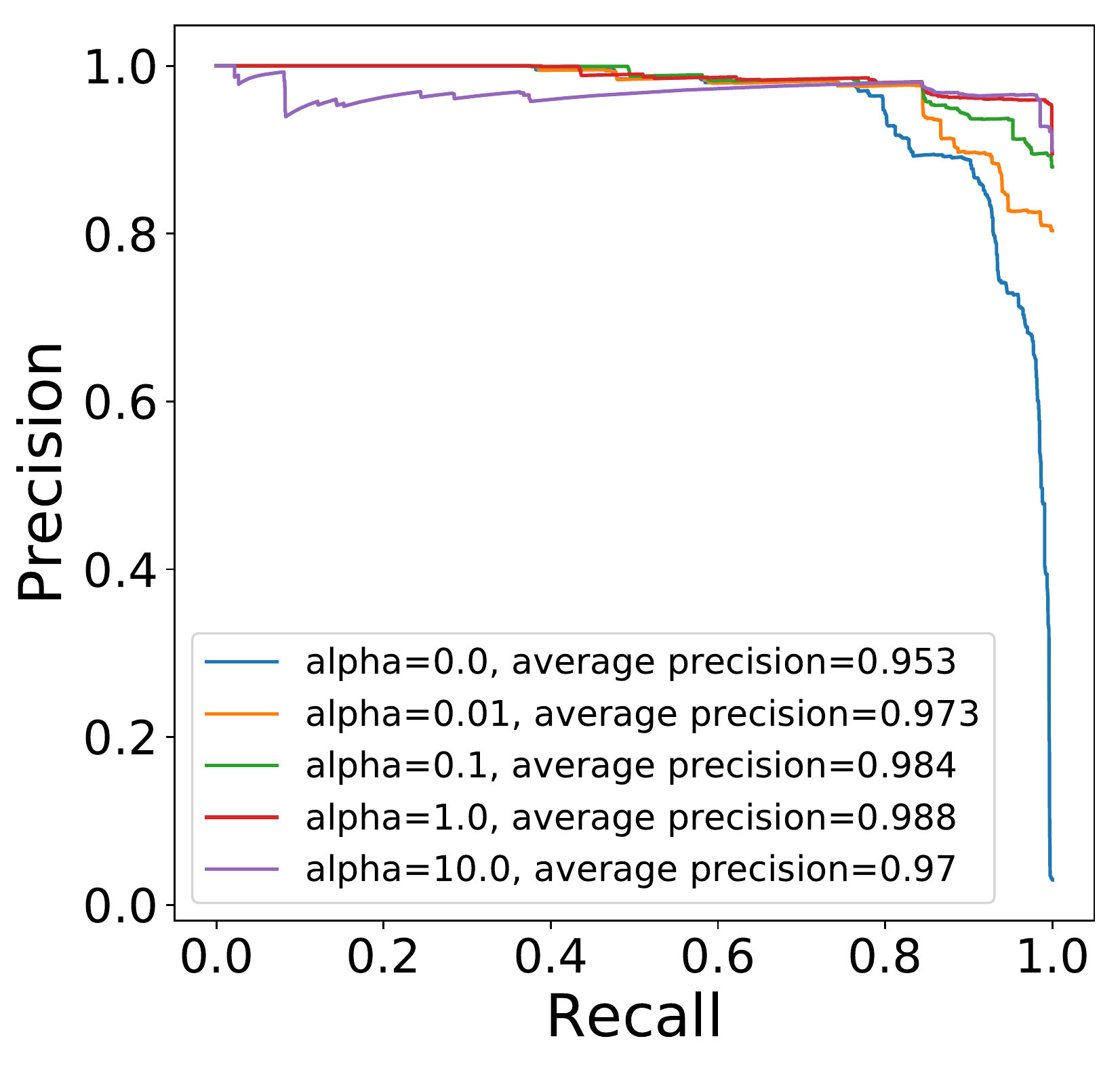} }}
\subfloat[Precision-Recall curves with \# of layers.\label{fig:layer}]{{\includegraphics[width=0.3\textwidth]{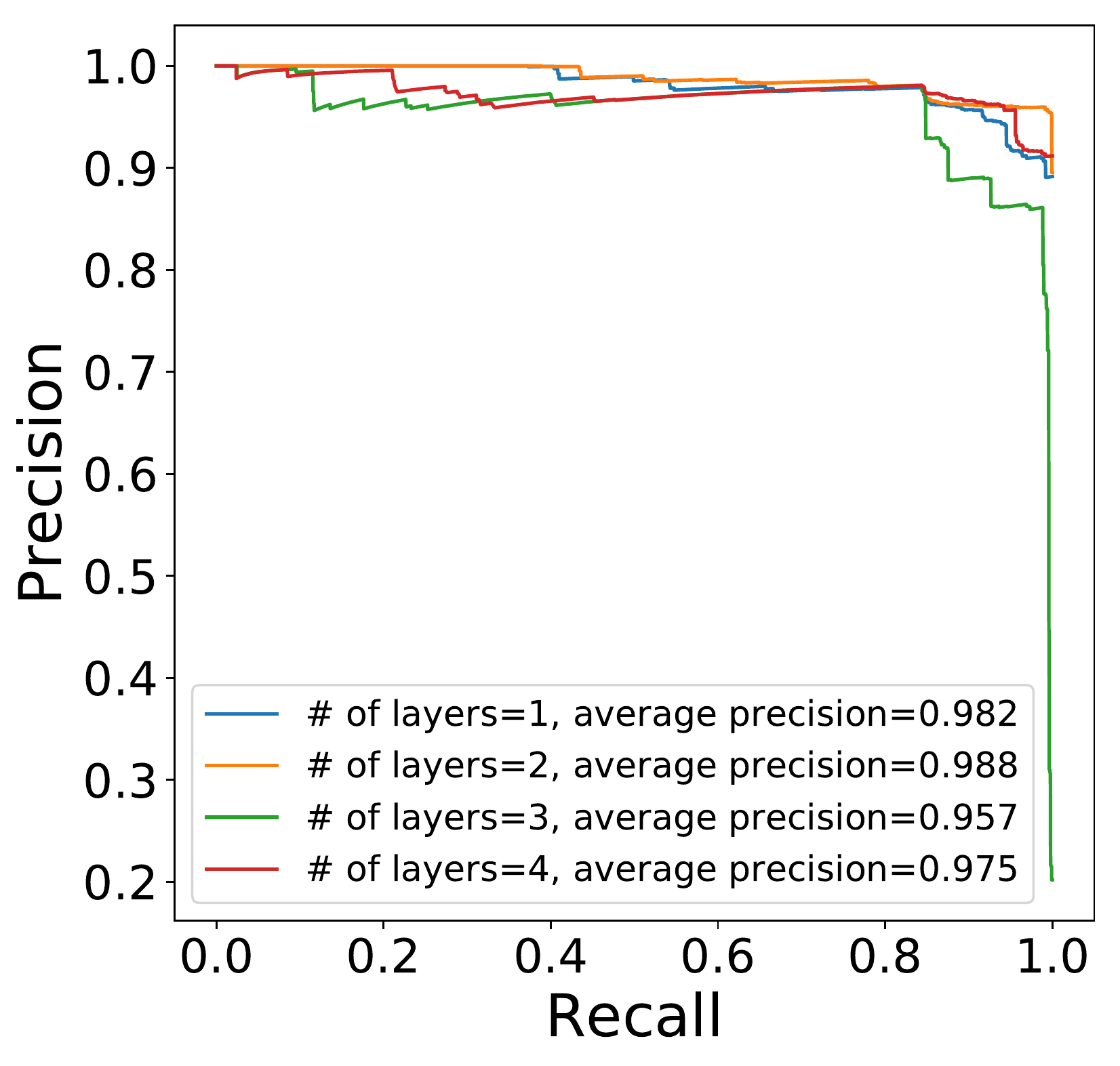} }}
\subfloat[Precision-Recall curves with different RNN cell types.\label{fig:cell}]{{\includegraphics[width=0.3\textwidth]{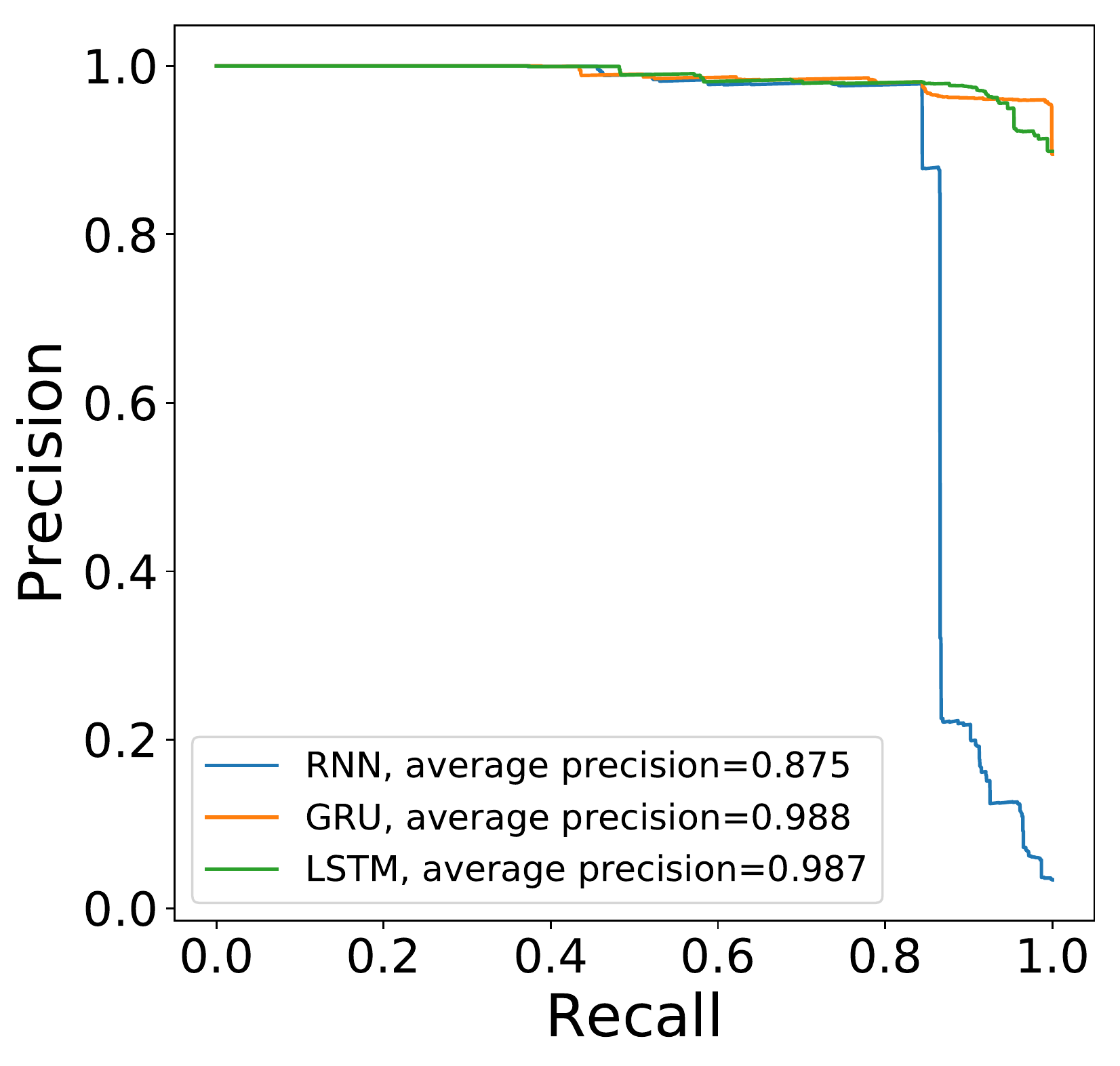} }}
\caption{Validation Precision-Recall curves on HDFS dataset.}
\label{fig:para}
\vspace{-3pt}
\end{figure*}

\begin{figure*}
\centering
\subfloat[Showcase of the importance of local information.\label{fig:case_local}]{{\includegraphics[width=0.43\textwidth]{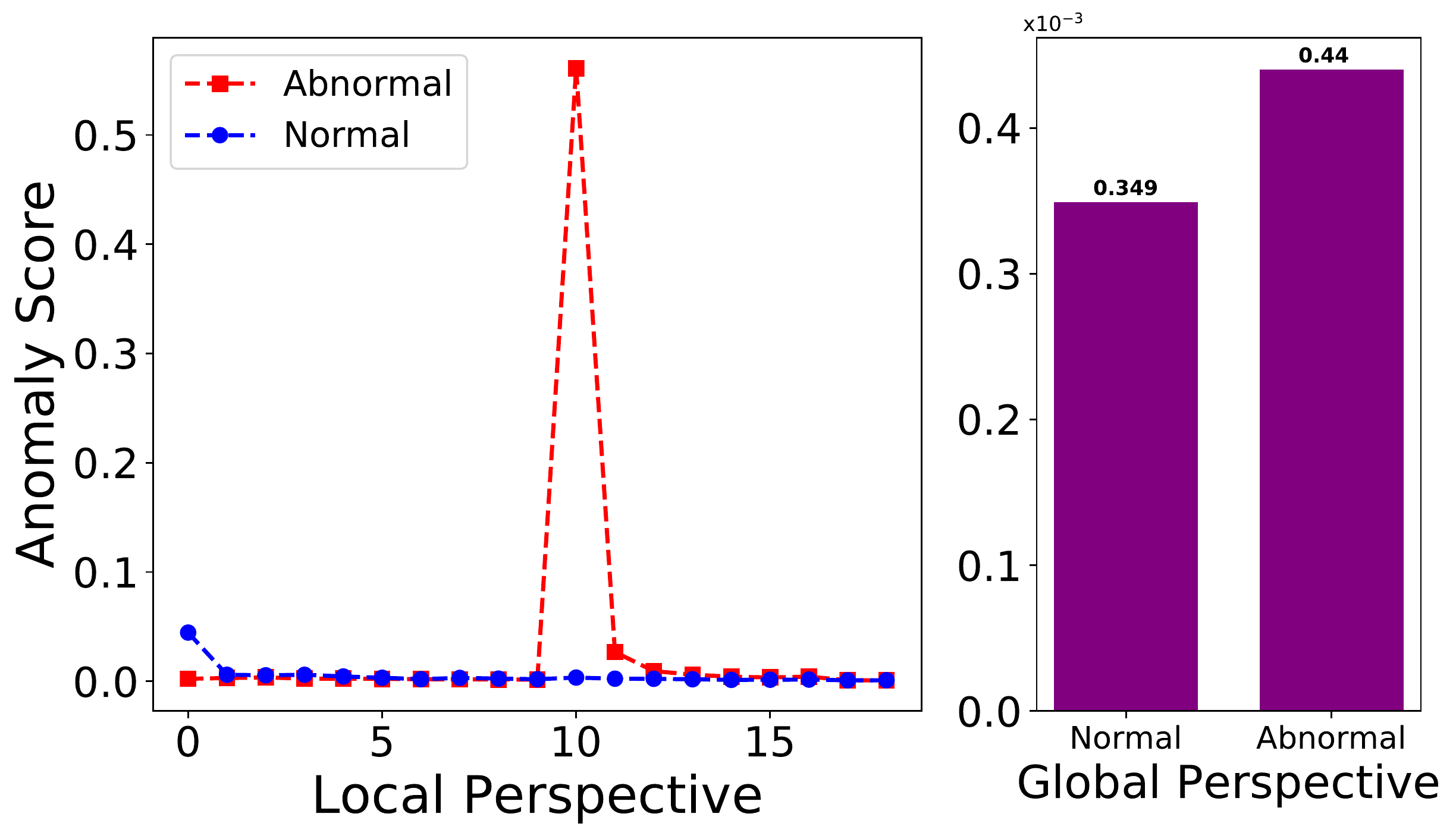}}}
\subfloat[Showcase of the importance of global information.\label{fig:case_global}]{{\includegraphics[width=0.45\textwidth]{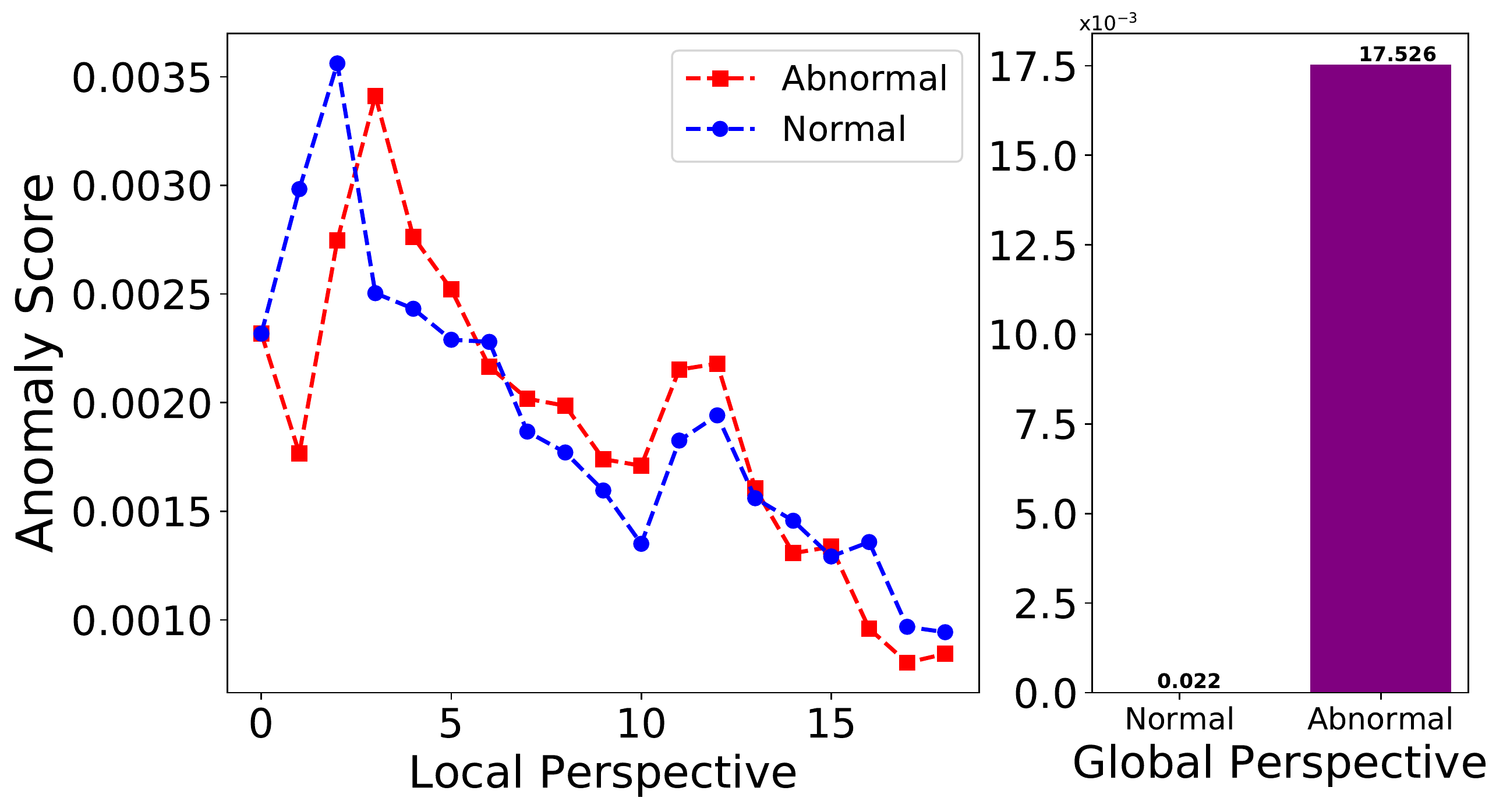} }}
\caption{The local and global anomaly scores of HDFS log key sequences.}
\label{fig:case}
\end{figure*}

\subsection{Performance Comparison}
The results of all methods on three datasets are shown in Table~\ref{tab:result}. From the table, we can make the following observations:
1) On most datasets, OC-SVM has the worst performance. We argue that this is because OC-SVM is highly dependent on feature qualities. Unlike dense data where OC-SVM generally performs very well, it is very hard to extract meaningful features from discrete event sequences. 2) IM and DeepLog outperform PCA significantly in most of the evaluation metrics. This is expected as both IM and DeepLog are designed specifically for anomaly detection for log data. 3) IM and DeepLog generally have comparable results in terms of F-1 score. It is interesting to observe that IM always achieves very impressive Recall while DeepLog is better in Precision. We argue that this difference could be caused by the fact that DeepLog focuses much on the local subsequence information while IM always concentrates on global sequence information. 4) The performance of DeepSVDD varies significantly in different datasets. This is because it only focuses on global information in the sequences. Thus, it can do very well with the datasets where local information is not important and becomes much worse otherwise. 5) On all the datasets, the proposed framework OC4Seq has achieved the best F-1 scores. In addition, when comparing to the second-best method, the performance gain brought by OC4Seq is significant. In terms of Precision, OC4Seq still achieved the highest value in most cases. For Recall, OC4Seq was only slightly outperformed by IM which has much lower precision scores. 6) All of the methods performed much worse on BGL datasets than the other two datasets. This is because BGL involves much more log keys that can make the task extremely difficult. Moreover, it is interesting to note that the DeepLog method becomes especially ineffective as it heavily relies on the next key prediction which is very difficult when log keys space becomes large.  In this challenging case, the improvement from OC4seq over other baselines becomes even more remarkable.

As a summary, from the experimental results on three datasets, our proposed framework OC4Seq demonstrates its superior performance over the representative baselines. We argue this is because OC4Seq can capture sequential information from both local subsequence and whole sequence and it directly optimizes the anomaly detection objective. Next, we design further experiments to gain deeper understandings of OC4Seq. 

\subsection{Parameter Analysis}
In this subsection, we analyze key hyper-parameters and components of OC4Seq. We only report the performance on the HDFS validation dataset as we have similar observations on the others. Moreover, the performance is evaluated by the Precision-Recall curve as it eliminates the need to choose a specific anomaly threshold and very suitable for datasets with imbalanced label distribution. We also report the area under the Precision-Recall curve (average precision) where the higher the value, the better the performance.

We firstly vary the value of $\alpha$ from $\{0,0.01,0.1,1,10\}$, which controls the contribution from local subsequence information. The results are shown in Figure~\ref{fig:alpha}. From the figure, we can see that with the increase of $\alpha$, the performance firstly increases and then decreases. The initial increase demonstrates the importance to incorporate local subsequence information while the latter decrease suggests that the global sequential information is also very essential and should not be overwhelmed by local information. 

Next, we vary the number of RNN layers from $\{1, 2, 3, 4\}$, and the results are shown in Figure~\ref{fig:layer}. These results suggest that more layers does not necessarily lead to better performance as it may cause other issues such as overfitting. Thus, it is important to select a proper value through the validation process.

Finally, to investigate how different types of RNN cell affect anomaly detection performance, we experienced on popular cells, \textit{i.e.}, RNN (Vanilla), GRU, LSTM. The results are shown in Figure~\ref{fig:cell}. From the figure, it is easily seen that the LSTM and GRU cells have very similar performance while vanilla RNN achieves much worse results. These results are consistent with previous work~\cite{chung2014empirical}. Due to its simpler structure, we choose GRU in OC4Seq.

\subsection{Case Study}
In this subsection, to further understand how the local and global information contribute to anomaly detection, we conduct case studies involving two pairs of representative log key sequences from HDFS dataset. Specifically, for a sequence, we use the trained model to calculate the anomaly scores of each subsequence and the whole sequence. The higher the anomaly score is, the more likely the sequence is abnormal. The results are shown in Figure~\ref{fig:case}. 

In Figure~\ref{fig:case_local}, we show the first pair of normal and abnormal sequences. The left panel uses dot lines to demonstrate the anomalous scores for local subsequences. Each dot denotes one anomalous score (y-axis) of $x^{th}$ subsequence (x-axis). The right panel uses the bar to show the anomalous score for the whole sequence (global information). From the figure, we observe that the two sequences have comparable anomalous scores for the whole sequence. Thus, it is very difficult to detect the abnormal one purely from the global perspective. However, from the local perspective, we can see that the $10^{th}$ subsequence of the abnormal sequence has a very high anomalous score while the anomalous scores of the subsequence of the normal sequence are all very low. Therefore, in this case, the local information plays a very important role in detecting anomalies. 

In Figure~\ref{fig:case_global}, the anomalous scores of the second pair of sequences are shown. Unlike the previous case, the second pair of sequences has very similar anomalous scores for local subsequences. This makes it hard to detect anomaly from the local perspective. However, the abnormal sequence has a significantly higher anomalous score from the global perspective than the normal one. Therefore, the global information contributes a lot to detecting the anomaly in this case. From the two cases, we further illustrate the importance of combining both local and global information in a sequence for anomaly detection.

\subsection{Visualization of Normal and Abnormal Sequences}
\begin{figure}
  \centering
  \includegraphics[scale=0.35]{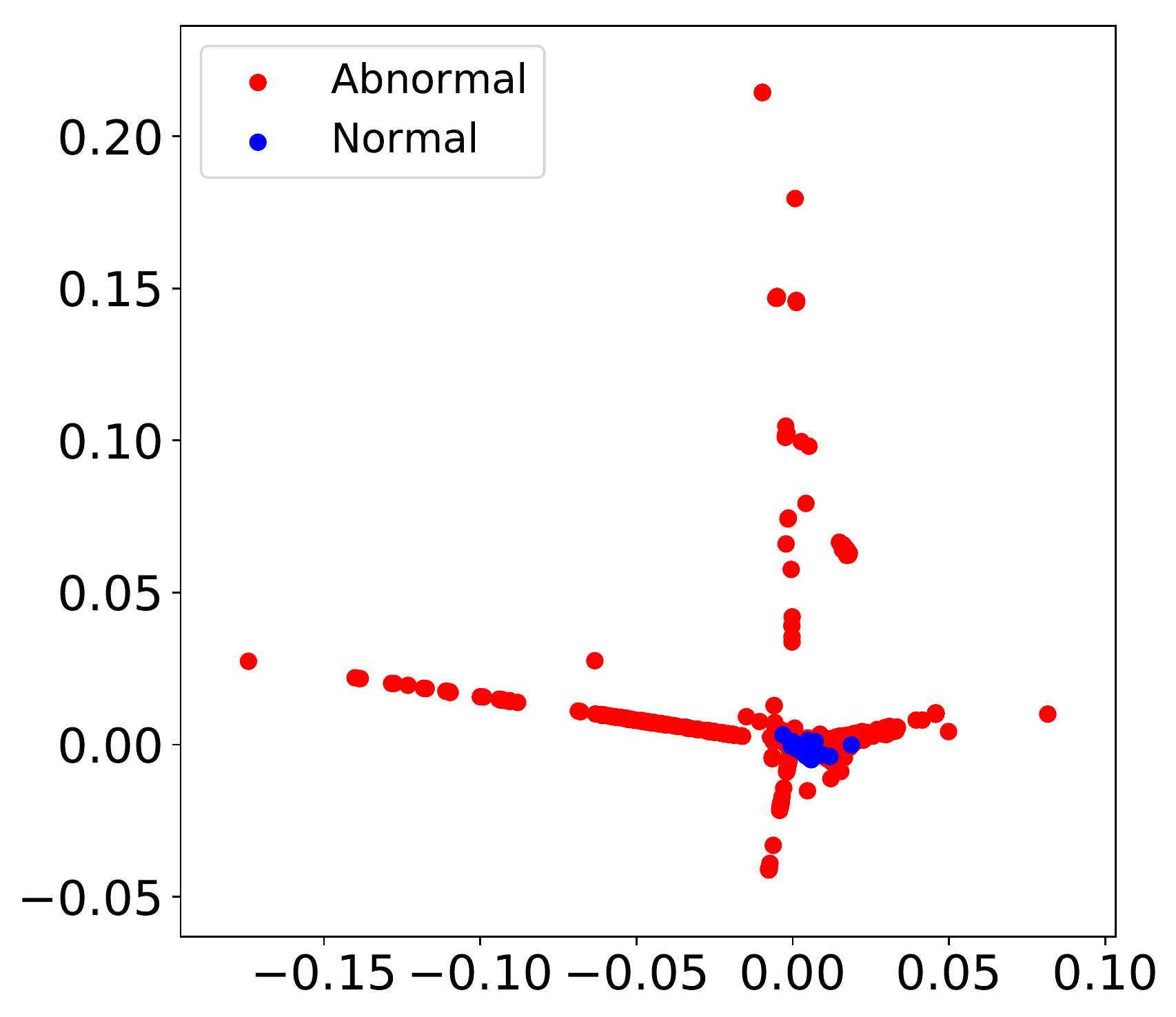}
  \caption{The visualization of the HDFS dataset. Abnormal and normal data are denoted by red and blue dots, respectively. It can be seen that normal sequences cluster in a small region while abnormal sequences spread over the latent space.}
  \label{fig:vis}
\end{figure}

To gain insights of the one-class classifier objective function, we project the global representations of both normal and abnormal sequences in the HDFS validation set to a two-dimensional space by Local Linear Embedding techniques~\cite{donoho2003hessian}. The visualization of the two-dimensional space is shown in Figure~\ref{fig:vis}. We observe that the normal sequences generally cluster together and lie in a very small region. On the other hand, the abnormal sequences spread all over the place. Therefore, the visualization clearly shows that by directly minimizing the anomalous score, which measures the distance between a normal data point and a center point, the one-class classifier objective function is very effective to guide the model to separate the abnormal and normal sequences in the latent space.

\section{Related Work}
In this section, we briefly review the related work on discrete event sequence anomaly detection and one-class classifier.

{\bf Anomaly Detection for Discrete Event Sequence:} There have been many research papers studying event sequence anomaly detection. One typical traditional anomaly detection method is similarity-based approach~\cite{budalakoti2006anomaly,budalakoti2008anomaly,boriah2008similarity}, where the similarity between a test sequence and training sequences is calculated and the anomalies are detected based on similarity scores. One major issue with similarity based methods is that there is a lack of intrinsic measurement of similarity between discrete event sequences and the effectiveness of the model can be largely affected by the choice of the similarity measurement. Other popular traditional methods include Markovian techniques and Hidden Markov Model based ones~\cite{cho2003efficient,ye2000markov,tan2008hidden,yamanishi2005dynamic}. However, the capacities of these models are too small to capture the complex long-term dependencies in the sequences. More recently, neural network based methods~\cite{du2017deeplog,hundman2018detecting,brown2018recurrent,tuor2018recurrent} have been proposed for anomaly detection and achieved great success due to its strong representation learning ability and large modeling capacities.

{\bf One-Class Classifier for Anomaly Detection:} 
One-class classifiers focus on learning the properties of objects from a single class and are trained to detect objects that are from any other classes. Therefore, they are very suitable for anomaly detection tasks and have gained great attention~\cite{perdisci2006using,wang2004anomaly}. The two most successful one-class classifiers are OC-SVM and SVDD~\cite{scholkopf2001estimating, tax2004support} that are based on traditional kernel tricks. Only recently, deep neural network based one-class classifier is proposed by Ruff {\it et al.}~\cite{ruff2018deep}. Inspired by SVDD~\cite{tax2004support}, the authors designed a novel one-class classification objective that has very nice theoretical properties and can effectively train CNNs for image anomaly detection tasks. Considering the success of one-class classifiers, we also build our framework upon similar objective functions. However, unlike previous works that mainly focus on dense data such as images, we conduct pioneering research of building one-class recurrent neural networks for discrete event sequences, which have very unique challenges comparing to the dense data.

\section{Conclusion}

In this paper, we propose OC4Seq, an end-to-end one-class recurrent neural network for discrete event sequence detection. OC4Seq can deal with multi-scale sequential dependencies and detect anomalies from both local and global perspectives. Specifically, OC4Seq incorporates an effective anomaly detection objective that can guide the learning process of sequence models. The trained multi-scale sequence model in OC4Seq explicitly maps the local subsequence and whole sequence into different latent spaces, where the normal data points are enclosed by hyperspheres with minimum volume. The proposed OC4Seq has consistently shown superior performance than representative baselines in extensive experiments on three benchmark datasets. In addition, through parameter analysis and case studies, the importance of capturing multi-scale sequential dependencies for discrete event sequence anomaly detection has been well demonstrated. Moreover, the visualization of the sequence representations qualitatively suggests the effectiveness of anomaly detection objectives.

%
% The next two lines define the bibliography style to be used, and the bibliography file.
\bibliographystyle{ACM-Reference-Format}
\balance
\bibliography{zhiwei}

\end{document}